\documentclass[letterpaper, 10 pt, conference]{ieeeconf}  % Comment this line out if you need a4paper

\IEEEoverridecommandlockouts                              % This command is only needed if 
\title{\LARGE \bf
Long-Short Term Spatiotemporal Tensor Prediction \\for Passenger Flow Profile
}

\author{ Ziyue Li $^{1}$, Hao Yan $^{\dagger 2}$, Chen Zhang $^{\dagger 3}$, Fugee Tsung $^{4}$ % <-this % stops a space
\thanks{*We show our great appreciation to the Metro Corporation for sharing this passenger flow data, and in the protection of privacy, all data has been desensitized.}% <-this % stops a space
\thanks{$\dagger$ Equal Contribution}
\thanks{$^{1}$Ziyue Li is now with Department of Industrial Engineering and Decision Analytics, Hong Kong University of Science and Technology, Hong Kong. 
        {\tt\small e-mail: zlibn@connect.ust.hk}}%
\thanks{$^{2}$Hao Yan is an Assistant Professor in School of Computing, Informatics, and Decision Systems Engineering, Arizona State University, 699 S Mill Ave, Tempe, AZ 85281.
        {\tt\small email: haoyan@asu.edu}}%
\thanks{$^{3}$Chen Zhang is an Assistant Professor in Industrial Engineering, Tsinghua University, Beijing, China.
        {\tt\small email: zhangchen01@tsinghua.edu.cn}}%
\thanks{$^{4}$Fugee Tsung is Chair  Professor in Department of Industrial Engineering and Decision Analytics, Hong Kong University of Science and Technology, Hong Kong.
        {\tt\small email: season@ust.hk}}%
}

\usepackage[english]{babel} %%% 'french', 'german', 'spanish', 'danish', etc.
\usepackage{amssymb}
\usepackage{amsmath}
\usepackage{txfonts}
\usepackage{mathdots}
\usepackage[classicReIm]{kpfonts}
\usepackage{graphicx}

\usepackage{lipsum} % just for the example
\usepackage[caption=false]{subfig}

\usepackage{algorithm,algcompatible,amsmath}
\algnewcommand\INPUT{\item[\textbf{Input:}]}%
\algnewcommand\OUTPUT{\item[\textbf{Output:}]}%

\usepackage{stmaryrd}

\usepackage{cleveref}

\begin{document}

\maketitle
\thispagestyle{empty}
\pagestyle{empty}

%%%%%%%%%%%%%%%%%%%%%%%%%%%%%%%%%%%%%%%%%%%%%%%%%%%%%%%%%%%%%%%%%%%%%%%%%%%%%%%%
\begin{abstract}

Spatiotemporal data is very common in many applications, such as manufacturing systems and transportation systems. It is typically difficult to be accurately predicted given intrinsic complex spatial and temporal correlations. Most of the existing methods based on various statistical models and regularization terms, fail to preserve innate features in data alongside their complex correlations. In this paper, we focus on a tensor-based prediction and propose several practical techniques to improve prediction. For long-term prediction specifically, we propose the "Tensor Decomposition + 2-Dimensional Auto-Regressive Moving Average (2D-ARMA)" model, and an effective way to update prediction real-time; For short-term prediction, we propose to conduct tensor completion based on tensor clustering to avoid oversimplifying and ensure accuracy. A case study based on the metro passenger flow data is conducted to demonstrate the improved performance.

\end{abstract}

%%%%%%%%%%%%%%%%%%%%%%%%%%%%%%%%%%%%%%%%%%%%%%%%%%%%%%%%%%%%%%%%%%%%%%%%%%%%%%%%
\section{INTRODUCTION}

Passenger flow data of an Urban Rapid Transit (URT) system is characterized as typical spatiotemporal data. Predicting passenger flow of a URT system has significant commercial value, such as automatical warning in advance when the system failure occurs for economic loss reduction. Based on the length of the prediction horizon, the task can be classified as short-term prediction (for several hours) and long-term prediction (for several days). Currently, the complicated spatial and temporal correlation structure hinders accurate prediction and further analysis \cite{bahadori2014fast}.

The goal of the spatiotemporal analysis is to capture various implicit spatial and temporal dependencies. In our URT case, for spatial dependency, the first aspect is the Law of Geography, where the passenger flow of a station is usually affected by its spatially adjacent neighbors. Fig. 1(a) plots the correlation structure of passenger flow data in stations along a specific URT line on a specific day. We can observe that neighboring stations are strongly correlated. The second aspect is the contextual similarity, where two stations sharing similar functions (business center, residential area or school, etc.) are more likely to have similar passenger flows. For temporal dependence, the future prediction is often correlated with historical observations in two different temporal scales such as weekly correlation and daily correlation as shown in Fig. 1(b, c). Weekly correlation refers to the passenger profile of today is correlated with the same day of previous weeks. Daily correlation refers to the the passenger profile of today is related to the pattern in the yesterday. 

Existing spatiotemporal forecasting methods are mainly based on geo-statistical models with regularization techniques. Different locations sharing similar features display common spatial correlations: Zhao \textit{et al}., used $l_{2,1}$-norm to achieve the relatedness of locations sharing the same keyword pool in social media analysis \cite{zhao2015multi}; Zhang \textit{et al}., also utilized $l_{2,1}$-norm to encourage all locations to select common features in citywide passenger flow prediction \cite{zhong2017spatiotemporal}. 
% Zhou \textit{et al}., use it to formulate  different patients sharing the same cognitive features when analyzing Alzheimer's Disease [3]

%To yield more accurate prediction of traffic flow, auxiliary information is also introduced to model. For example, Zhong \textit{\textit{et al}., }[4] consider additional features including traffic capacity, regional characteristics, event and weather, although this involves a heavy workload. Some domain-specific knowledge is required for specific cases. For the price spatiotemporal data, its spatial correlation also follows the Law of One Price policy [5].

\begin{figure}[t]
\centering
%\framebox{\parbox{3in}{We suggest that you use a text box to insert a graphic (which is ideally a 300 dpi TIFF or EPS file, with all fonts embedded) because, in an document, this method is somewhat more stable than directly inserting a picture.}}
\includegraphics[width=0.95\columnwidth]{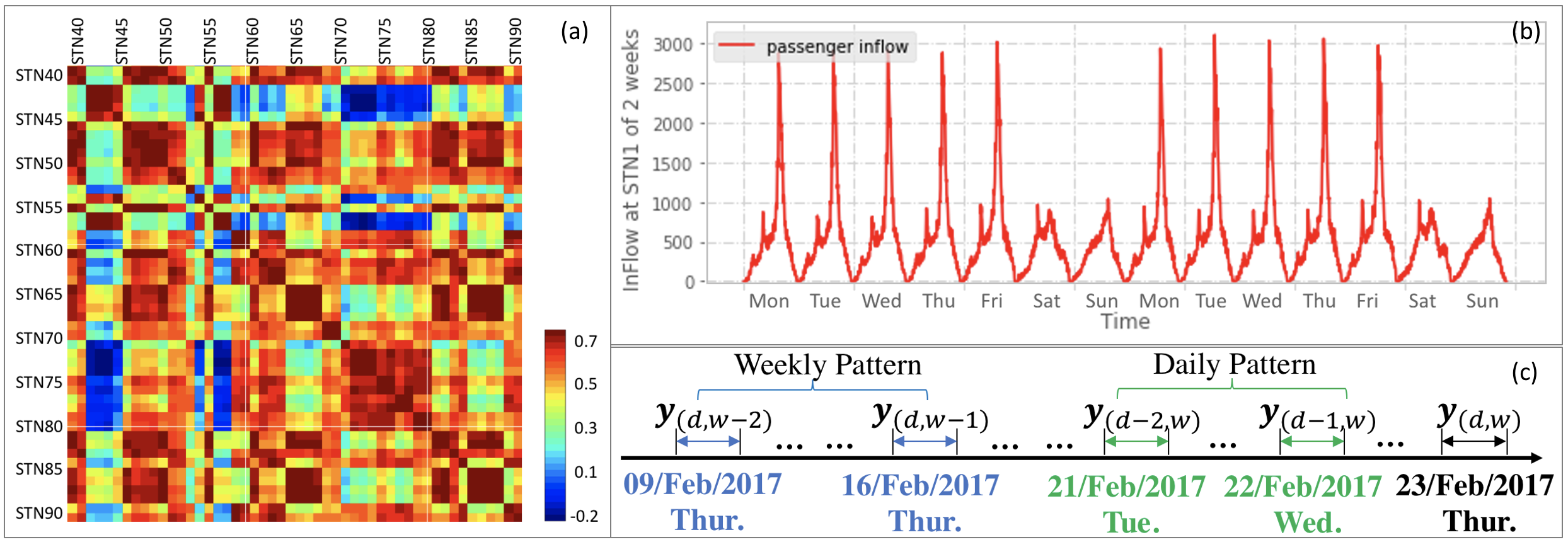} % Reduce the figure size so that it is slightly narrower than the column. Don't use precise values for figure width.This setup will avoid overfull boxes. 
\caption{(a) Spatial Correlation. (b) Inflow Profile of 2 weeks for STN1 (Station names are desensitized as station code ‘STN$\#$’). (c) Daily and Weekly Dependencies observed in Data.}
\label{fig1}
\end{figure}

To capture the temporal pattern, many traditional time-series models have been developed for traffic prediction, such as Holt-Winters forecasting \cite{tikunov2007traffic} and Auto-Regressive Integrated Moving Average (ARIMA) \cite{williams2003modeling}, which incorporate the regularization terms. For example, temporal smoothness is often enforced by penalizing the difference between two consecutive timestamps $\left\|\boldsymbol{w}^t-\boldsymbol{w}^{t-1}\right\|$ \cite{zhang2017spatiotemporal}. However, the linear parametric form with few lags imposes strong assumptions on the temporal correlation, which may lead to under-fitting. Most importantly, a traditional ARIMA is powerless to present the cross-dependency of the spatial and temporal dimension, which is severely overlooked in most of the above research. 

Another way to deal with the spatiotemporal prediction is the application of Neural Networks. Convolutional Neural Network (CNN) is a common way to capture spatial correlation by using a station’s neighbors to predict its future behavior. Recurrent Neural Network (RNN) and its variants (e.g., Long Short Term Memory networks) are sufficiently capable of modeling complex temporal correlation. Built upon CNN and LSTM, H. Yao \textit{et al}. \cite{yao2018deep} developed a spatio-temporal network to predict taxi demand of different regions. It adds a further layer to consider: semantic similarity of locations, meaning that locations sharing a similar functionality may have similar demand patterns. Recently, X. Geng \textit{et al}. \cite{geng2019spatiotemporal} incorporated Multi-Graph Convolution Network to interpret not only the Euclidean correlation among spatially adjacent regions but also involved Non-Euclidean adjacency, including function similarity and Connectivity, following by Gated Recurrent Neural Network (GRNN) to capture temporal dependency.
%However, the normal convolutional operation that is usually applied restricts the model to only process grid structure rather than general domains. Besides, LSTM and RNN also require iterative training, which introduces error accumulation by steps. 
However, it not only requires costly computational resources to train these deep learning models but also demands a very large sample size for training. 

In this research, we aim to develop a computationally efficient and robust  way to deal with the spatiotemporal prediction problem. To achieve this, we propose to use the tensor representation of the spatio-temporal data, which is proven as an efficient way to represent spatiotemporal data due to its sufficient capacity to capture inter-dependencies along multiple dimensions. In particular, there are two conventional ways to conduct spatiotemporal prediction by tensor, and we categorize them as 2-step tensor prediction and 1-step tensor prediction in the following.

\paragraph{2-step Tensor Decomposition and Time Series Modeling for Spatio-temporal prediction}

These methods combine tensor decomposition combined with traditional time-series prediction models \cite{dunlavy2011temporal,guo2017novel,ren2017efficient}, and it is suitable for long-term prediction.

Tensor Decomposition can be considered as a high-dimensional version of matrix singular value decomposition \cite{kolda2009tensor}. Two specific forms of tensor decomposition are usually adopted in tensor analysis: CANDECOMP/PARAFAC (CP) that decomposes a tensor as a sum of rank-one tensors, and Tucker Decomposition that decomposes a tensor into a core tensor multiplied by each mode matrix.

Moreover, the 2-step tensor prediction initially conducts Tensor Decomposition to obtain the temporal mode matrix along time dimension, and then exploits time-series model on the temporal mode. Holt-Winters forecasting \cite{dunlavy2011temporal}, Auto-Regressive Integrated Moving Average (ARIMA) and Support Vector Regression (SVR) \cite{ren2017efficient} have been exploited. 

However, existing 2-step tensor prediction methods are not capable of capturing all the weekly and daily patterns mentioned above. Take ARIMA or AR model for example. If the weekly pattern is desired, then time-lag should be set large to include past few weeks at least, which highly complicates the model. To address this, we proposed to first reshape the daily profile into a matrix (in the form of ${\mathbb{R}}^{day\times week}$) and then apply a 2-Dimensional Auto-Regressive Moving Average (2D-ARMA) model to capture all those daily and weekly patterns. We name it as "2-step 2D-ARMA" tensor prediction. Finally, real-time prediction update when new data arrive is another challenge. To this end, we also proposed a Lean Dynamic Updating method. %introduces plenty of time-series model, which increase computational burden.

\paragraph{1-step Tensor Prediction based on tensor completion}
This is based on tensor completion \cite{tan2016short,ran2016tensor, luan2019prediction}, and it is suitable for short-term prediction (prediction horizon as several hours ahead). Tensor Completion is originally designated for tensor random missing data imputation. H. Tan \textit{et al}., used tensor completion first time for traffic volume prediction \cite{tan2016short,chen2019bayesian}, which treated future data as missing data to be estimated.

%I think tensor rank problem is not a concern in our paper, and we are not offering a solution for it.
%However, either for two-step or one-step methods based on tensor decomposition or tensor completion, the challenge rests as follows: to determine the rank of a specific given tensor is an NP-hard problem [16]. For example,  the rank of a particular Tensor ${\boldsymbol{\mathcal{X}}}^{\mathrm{9\times 9\times 9}}$ can be only determined to be bounded in $ 18 \leqslant rank \leqslant 23$, let alone the real high-dimensional tensor. Currently, there is a large body of research focusing on solutions to this problem.  For tensor decomposition, researchers added orthogonal constraints to ease the complexity [14]. 
For tensor completion, Low-Rank Tensor-Completion (LRTC) method has prevailed. One of the most common technique is adding a nuclear-norm on tensor's rank \cite{kressner2014low,yokota2016smooth,liu2012tensor}; Q. Shi \textit{et al}., tried $l_1$-norm on CP weight vector \cite{shi2017tensor}; Q. Zhao \textit{et al}., also proposed a CP-based Bayesian Hierarchical Probabilistic model, assuming that all mode matrices were generated from higher-level latent distribution, with sparsity-inducing prior to low-rank \cite{zhao2015bayesian}.

However, 1-step tensor prediction based on LRTC is prone to oversimplify the model, which results in loss of prediction accuracy. To solve this problem, we propose to first cluster spatiotemporal data and then conduct LRTC within data from the same cluster. To this end, Tensor clustering method will also be studied.

In this paper, we will focus on the improving  both 1-step and 2-step state-of-the-art tensor prediction methods, and provide practical and effective techniques to improve the prediction performance correspondingly. In summary, this paper makes the following contributions: 

\begin{itemize}
\item  We improve both 2-step and 1-step tensor prediction for URT passenger flow prediction. For example, for 2-step tensor prediction, we propose a 2-step 2D-ARMA tensor prediction model. For 1-step tensor prediction, we improve the LRTC by conducting it together with a tensor cluster algorithm.

\item  Furthermore, for both methods, we also propose how to dynamically apply the proposed method online. For example, we propose a Lean Dynamic Updating method for tensor decomposition to update the previous prediction real-time;

\end{itemize}

\section{Preliminaries}

We first review the preliminaries and backgrounds about tensor decomposition methods and tensor completion.

\subsection{Notations and Operations}

Throughout this exposition, scalars are denoted in italics, e.g. $n$; vectors by lowercase italic letters in bold face, e.g. $\boldsymbol{u}$; and matrices by uppercase boldface letters, e.g. $\boldsymbol{\mathrm{U}}$; High dimensional data, tensor by boldface script capital $\boldsymbol{\mathcal{X}}$.

\subsection{Tensor Decomposition and Completion}

We will introduce the basic knowledge of CP, Tucker Decomposition and Tensor Completion here.

\paragraph{CP Decomposition}

A tensor $\boldsymbol{\mathcal{X}}\in {\mathbb{R}}^{I_1\times I_2\times \cdots \times I_K}$ is represented as the weighted summation of a set of rank-one tensors: 
\begin{equation} \label{GrindEQ__1_} 
\begin{split}
\boldsymbol{\mathcal{X}} &=\sum^R_{r=1}{{\lambda}_r{{\boldsymbol{u}}_r}^{\left(1\right)}\circ {{\boldsymbol{u}}_r}^{\left(2\right)}\circ \cdots \circ{{\boldsymbol{u}}_r}^{\left(K\right)}}\\
&=\llbracket \boldsymbol{\lambda};\mathbf{U}^{(1)},\mathbf{U}^{(2)},...,\mathbf{U}^{(K)}\rrbracket ,
\end{split}
\end{equation} 
 where each ${{\boldsymbol{u}}_r}^{\left(k\right)}(k=1,\dots,K)$ is a unit vector, and $\circ $ is the outer product. ${\boldsymbol{\mathrm{U}}}^{(k)}\in {\mathbb{R}}^{I_K\times R}(k=1,\dots,K)$ is the mode matrix of Dimension-$k$ and $R$ is the rank of CP decomposition. $\boldsymbol{\lambda}=[\lambda_1,\cdots,\lambda_R]$ is the score vector.
 
 \paragraph{Bayesian Low-rank Tensor Decomposition}

Consider ${\boldsymbol{\mathcal{Y}}}^{I_1\times I_2\times \cdots \times I_K}$ is a noisy observation of  tensor $\boldsymbol{\mathcal{X}}$, i.e., $\boldsymbol{\mathcal{Y}}=\boldsymbol{\mathcal{X}}\mathrm{+}\mathrm{\varepsilonup }$, and the noise is assumed to be an i.i.d Gaussian distribution $\mathrm{\varepsilonup }\mathrm{\sim }\prod_{i_1,\dots ,i_K}{\mathcal{N}(0,{\tau }^{-1})}$. $\boldsymbol{\mathcal{X}}$ is generated by CP model, with weight absorbed inside of mode matrices.

Mode-$k$ factor matrix ${\boldsymbol{\mathrm{U}}}^{\mathrm{(}k\mathrm{)}}$ can be denoted by row wise or column wise vectors ${\boldsymbol{\mathrm{U}}}^{(k)}={\left[{\boldsymbol{u}}^{\left(k\right)}_{i_k},\dots ,{\boldsymbol{u}}^{\left(k\right)}_{i_k},\dots ,{\boldsymbol{u}}^{\left(k\right)}_{i_k}\right]}^T=\left[{{\boldsymbol{u}}_1}^{\left(k\right)},\dots ,\ {{\boldsymbol{u}}_r}^{\left(k\right)},\dots ,{{\boldsymbol{u}}_R}^{\left(k\right)}\right]$.

The generative model based on Bayesian probabilistic structure \cite{zhao2015bayesian} is shown in Fig. 2, and is specified as:
\begin{itemize}
\item $\boldsymbol{\lambda}$ and $\tau $ are generated by:
\begin{equation} \label{GrindEQ__6_}
\begin{split}
P(\boldsymbol{\lambda}) &
=\prod^R_{r=1}{Ga\left(\left.{\lambda }_r\right|c^r_0,d^r_0\right)},\\
P\left(\tau \right)&
=Ga\left(\left.\tau \right|a_0,b_0\right) .
\end{split}
\end{equation} 

\item ${\boldsymbol{\mathrm{U}}}^{(k)}$ (given $\boldsymbol{\lambda }$) is generated by: 
\begin{equation} \label{GrindEQ__7_} 
P\left(\left.{\boldsymbol{\mathrm{U}}}^{\left(k\right)}\right|\boldsymbol{\lambda }\right)=\prod^{I_k}_{i_k=1}{N\left(\left.{\boldsymbol{u}}^{\left(k\right)}_{i_k}\right|0,\ {\mathrm{\Lambda }}^{-1}\right)},\mathrm{\Lambda }\mathrm{=diag}(\boldsymbol{\lambda }).
\end{equation} 
%Since the priors are shared across $K$ latent matrices, the same sparsity pattern can be obtained, which contributes to the minimal number of rank-one tensors and achieves low-rank objectives.

\begin{figure}[t]
\centering
%\framebox{\parbox{3in}{We suggest that you use a text box to insert a graphic (which is ideally a 300 dpi TIFF or EPS file, with all fonts embedded) because, in an document, this method is somewhat more stable than directly inserting a picture.}}
\includegraphics[width=0.95\columnwidth]{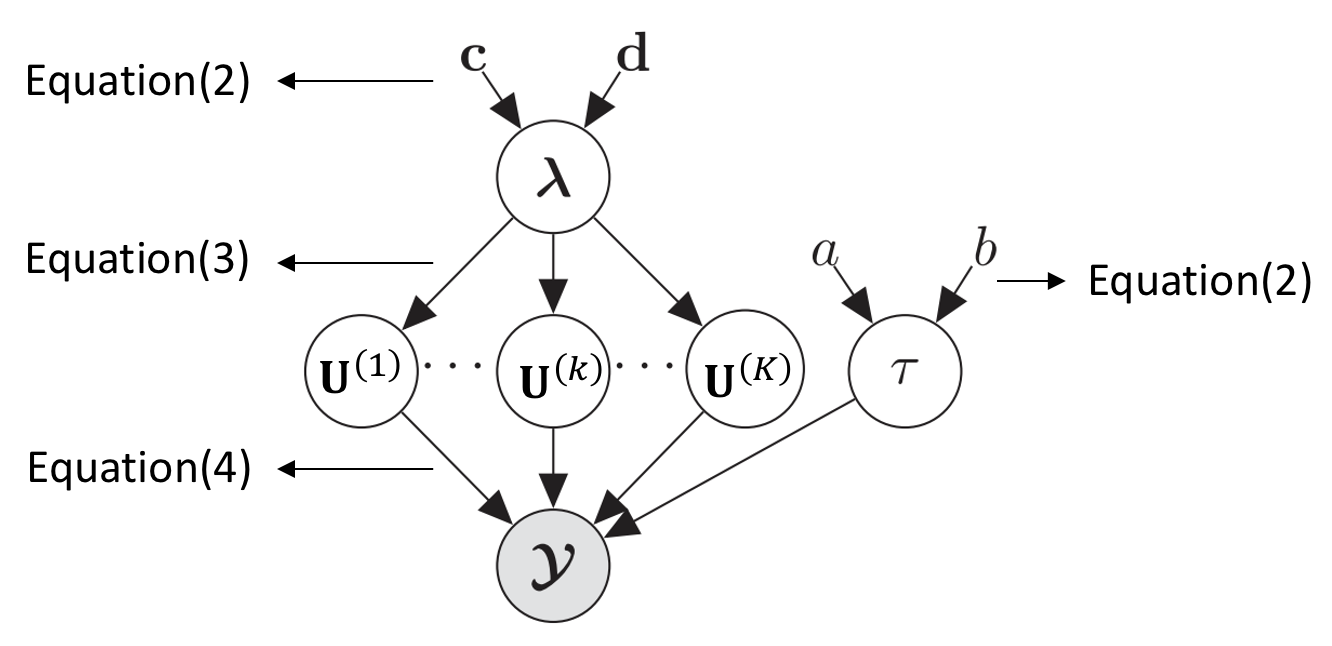} % Reduce the figure size so that it is slightly narrower than the column. Don't use precise values for figure width.This setup will avoid overfull boxes. 
\caption{Bayesian Probabilistic Structure}
\label{fig6}
\end{figure}
\item ${\boldsymbol{\mathcal{Y}}}_{\boldsymbol{\mathrm{\Omega}}}$ (given ${\left\{{\boldsymbol{\mathrm{U}}}^{(k)}\right\}}^K_{k=1}\mathrm{,\ }\tau $) is generated by: 
\[P\left(\left.{\boldsymbol{\mathcal{Y}}}_{\boldsymbol{\mathrm{\Omega}}}\right|{\left\{{\boldsymbol{\mathrm{U}}}^{\left(k\right)}\right\}}^K_{k=1}\mathrm{,\ }\tau \right)=\] 
\begin{equation} \label{GrindEQ__9_} 
\prod^{I_1}_{i_1=1}{\cdots \prod^{I_k}_{i_k=1}{N{\left(y_{i_1i_2\dots i_N}|\left\langle {\boldsymbol{u}}^{\left(1\right)}_{i_1},{\boldsymbol{u}}^{\left(2\right)}_{i_2},\cdots ,{\boldsymbol{u}}^{\left(K\right)}_{i_K}\right\rangle ,{\tau }^{-1}\right)}^{{\boldsymbol{O}}_{i_1i_2\dots i_N}}}}.
\end{equation} 
\end{itemize}

\paragraph{Tucker Decomposition}

It is commonly regarded as higher-order PCA, decomposing a tensor $\boldsymbol{\mathcal{X}}\in {\mathbb{R}}^{I_1\times I_2\times \cdots \times I_K}$ into  a core tensor multiplied by a matrix along each mode, i.e.,
\begin{equation} \label{GrindEQ__2_} 
 \boldsymbol{\mathcal{X}}=\boldsymbol{\mathcal{G}}{\times }_1{\boldsymbol{\mathrm{U}}}^{\mathrm{(1)}}{\mathrm{\times }}_{\mathrm{2}}{\boldsymbol{\mathrm{U}}}^{\mathrm{(2)}}\mathrm{\cdots }{\mathrm{\times }}_{\mathrm{K}}{\boldsymbol{\mathrm{U}}}^{\mathrm{(}K\mathrm{)}},
\end{equation} 
where $\boldsymbol{\mathcal{G}}\in {\mathbb{R}}^{J_1\times J_2\times \cdots \times J_K}$ is core tensor,  ${\mathbf{U}}^{\mathrm{(}k\mathrm{)}}\in {\mathbb{R}}^{I_k\times J_k}$, and $[J_1,J_2,\dots ,J_K]$ is the rank for Tucker decomposition.

\paragraph{Tensor Completion}
It is usually designed for random missing data imputation, like random pixel missing in image data \cite{zhao2015bayesian}. The basic tensor completion is formulated as following: 
\begin{equation} \label{GrindEQ__4_} 
{\mathop{\mathrm{min}}_{\boldsymbol{\mathcal{Y}}} \ \left\|{\boldsymbol{\mathcal{X}}}_{\boldsymbol{\mathrm{\Omega}}}-{\boldsymbol{\mathcal{Y}}}_{\boldsymbol{\mathrm{\Omega}}}\right\|+\alpha {\left\|\boldsymbol{\mathcal{Y}}\right\|}_*\ }, 
\end{equation} 
where $\boldsymbol{\mathcal{X}}$ is the incomplete input tensor, $\boldsymbol{\mathcal{Y}}$ is the completed output matrix, ${\left\| \cdot \right\|}_{*}$ is the nuclear norm to achieve low rank, and $\boldsymbol{\mathrm{\Omega}}$ is sampling set which denotes the indices of the observed elements of a tensor.

\section{Proposed Tensor Prediction Framework}
In this section, we aim to propose spatio-temporal prediction framework based on the tensor decomposition and tensor completion methods in Section II. 
For the short-term prediction, the effective prediction horizon is 2 hours ahead, which is the response time needed for URT company to take corresponding actions when abnormal passenger flow pattern happens. In the case of URT passenger flow, data can be represented as: ${\boldsymbol{\mathcal{X}}}^{L\times T\times P}$. $L$ is location standing for 120 stations in our dataset, $T$ is the time scope we are looking at (here is 1$^{st}$ Jan 2017 to 28$^{th}$ Feb 2017), $P$ is the 5-minute interval observations per day with 247 sensing points. Thus our data is ${\boldsymbol{\mathcal{X}}}^{120\times 59\times 247}$.

Our prediction problem can be formulated as: 
\begin{equation} \label{GrindEQ__3_} 
{\boldsymbol{\mathcal{X}}}^{L\times (T+\tau )\times P}=f({\boldsymbol{\mathcal{X}}}^{L\times T\times P}), 
\end{equation} 
where $\tau $ is the prediction horizon.

\subsection{2-step Tensor Long-Term Prediction: Combine Tensor Decomposition and 2D-ARMA}

2-step tensor prediction is popularly used in the past few years and the mechanism(shown as Fig. 3(a)) behind is designed as: 

\begin{itemize}
\item  Formulate data as a tensor form. In our case, it is ${\boldsymbol{\mathcal{X}}}^{L\times T\times P}$. Decompose it, and among the decomposed components we can find temporal mode matrix ${\boldsymbol{\mathrm{U}}}_T$.

\item  Use traditional time-series model to predict the incoming $\tau $ time's temporal model matrix ${\boldsymbol{\mathrm{U}}}_{T+\tau } = f({\boldsymbol{\mathrm{U}}}_T)$.

\item  Then substitute it back to decomposition structure to reconstruct the tensor ${\boldsymbol{\mathcal{X}}}^{L\times (T+\tau) \times P}$.
\end{itemize}

\begin{figure}[htp]
\subfloat[Tensor Decomposition + 2D ARMA Model]{%
  \includegraphics[clip,width=0.95\columnwidth]{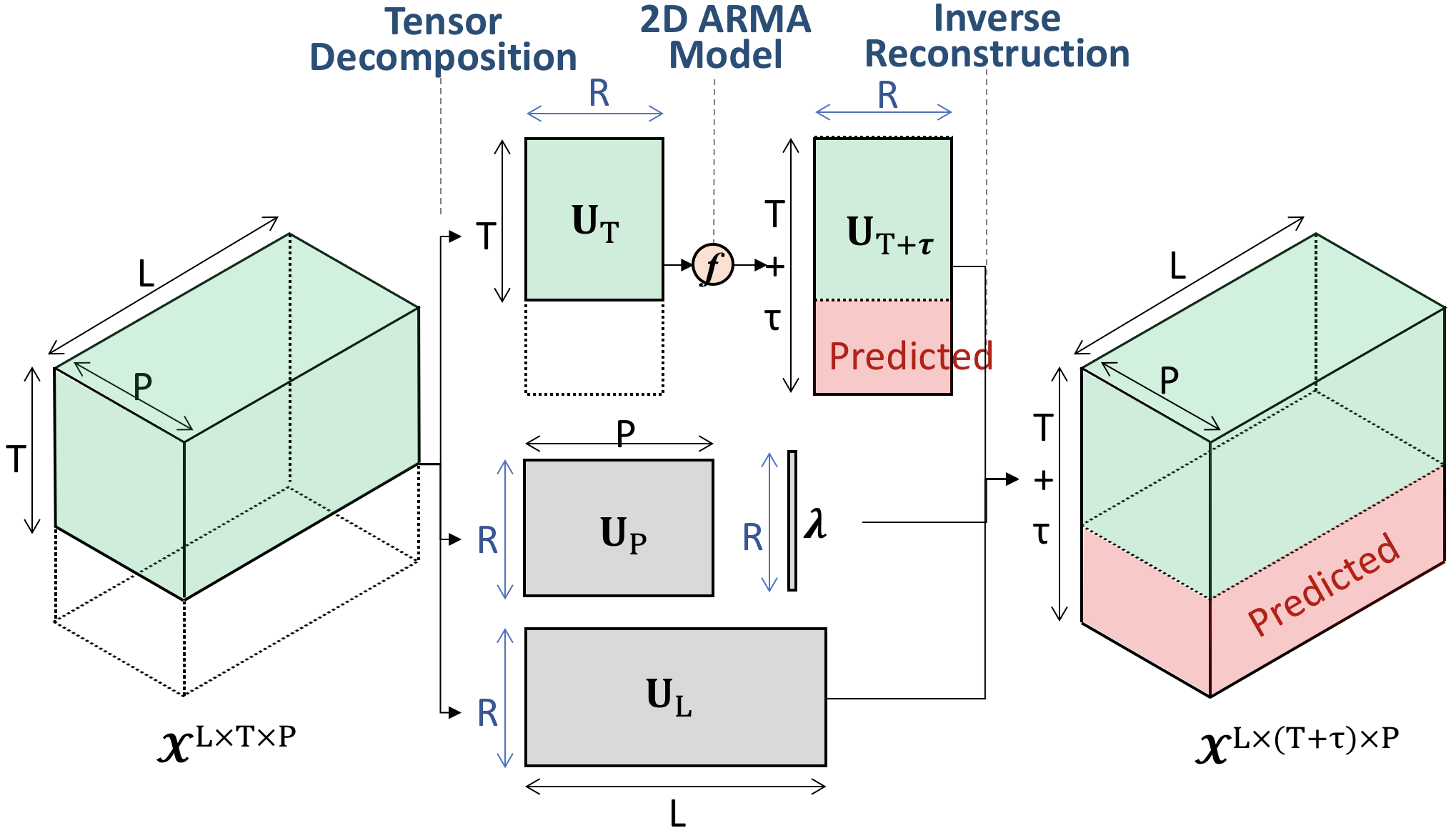}%
}

\subfloat[2D ARMA Model on Rank-r Vector of ${\boldsymbol{\mathrm{U}}}_T$]{%
  \includegraphics[clip,width=0.95\columnwidth]{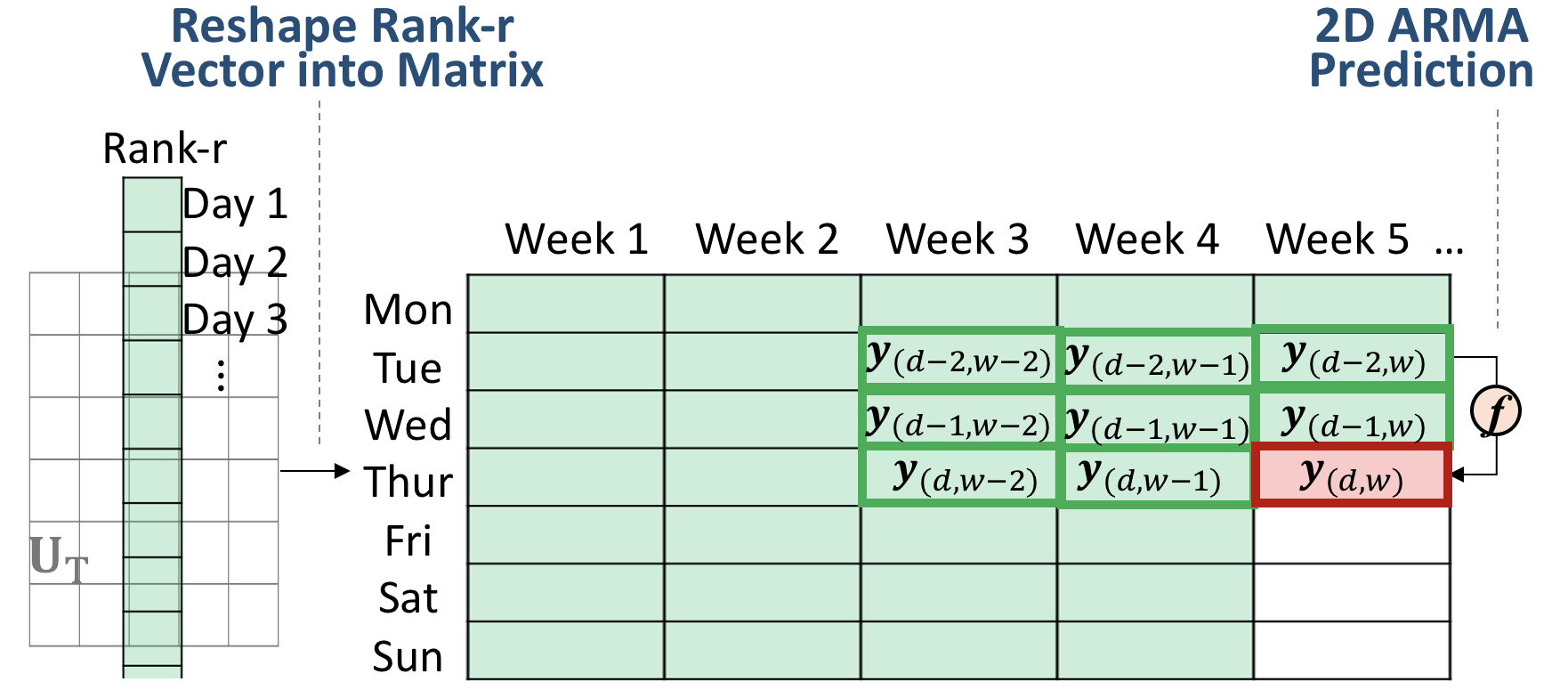}%
}

\caption{2-step 2D-ARMA Tensor Prediction}

\end{figure}

For time-series model, options include Holt Winters, ARIMA etc. However, as mentioned before, they are unable to capture the temporal pattern as shown in Fig. 1(c), especially the weekly pattern. This is because, to take the same day of past 2 weeks, at least 14 days time-lags should be involved into model, which leads to too much model complexity. 

Therefore, we proposed the adoption of 2D-ARMA model for the prediction step (shown in Fig. 3(b)), with following additional steps needed:

\begin{itemize}
\item  For the Rank-r vector of ${\boldsymbol{\mathrm{U}}}_T$, $r=1,2,..., R$, reshape it into 2D matrix, with each row representing one certain day of the week, each column representing the seven days in one week.
%and $D \times W \geqslant R$ and set the  $(D \times W - R)$ new generated elements as 0; 

\item  Train the 2D-ARMA model using ${\boldsymbol{\mathrm{U}}}_T$.
\item  After prediction, vectorize the matrix back to vector, and combine all the rank vectors to obtain ${\boldsymbol{\mathrm{U}}}_{T+\tau }$.
\end{itemize}

%The most significant issue facing 2-step tensor prediction is as follows: when predicting the temporal mode matrix, for time-series profile of each rank (each column in Fig. 3(a)), a specific time-series model (ARIMA model for instance) needs to be constructed. In Fig. 3(a), the rank of temporal mode is 50 for instance, so 50 ARIMA models need to be constructed and tuned, which creates challenges.

The matrix from Rank-r vector $\boldsymbol{u}_r$ of ${\boldsymbol{\mathrm{U}}}_T$  is represented as a 2D random field $v[d,w], d \in \mathbb{R}^{D=7}, w \in \mathbb{R}^{W}$. The 2D ARMA$(p_{1}, p_{2}, q_{1}, q_{2})$ model is defined for the $D \times W$ for the matrix $\boldsymbol{\mathrm{V}}=\{v[d,w]: 0 \leqslant d \leqslant D-1, 0 \leqslant w \leqslant W-1\}$ by the following equation:
\begin{equation} \label{GrindEQ__4_} 
\begin{split}
 v[d,w] 
 & + \mathop{\sum^{p_{1}}_{i=0}\sum^{p_{2}}_{j=0}}\limits_{(i,j) \neq (0,0)}a_{ij}v[d-p_{1}, w-p_{2}] \\
 & = \sum^{q_{1}}_{i=0}{\sum^{q_{2}}_{j=0}{b_{ij}\varepsilonup[d-p_{1}, w-p_{2}]}}.
\end{split}
\end{equation} 
$\{\varepsilonup[d, w]\}$ is a stationary white noise with variance $\sigma^2$, and the coefficients of $\{a_{ij}\}, \{b_{ij}\}$ are the parameter of the model. $(p_{1}, p_{2})$ and $(q_{1}, q_{2})$ are the time lags of $(d,w)$ for $v$ and $\varepsilonup$ respectively. The parameter estimation is explained in \cite{zielinski2010two}. The 2-step 2D-ARMA Tensor Prediction is summarized in Algorithm 1.

\begin{algorithm}
    \caption{2-step 2D-ARMA Tensor Prediction}
  \begin{algorithmic}[1]
    \INPUT ${\boldsymbol{\mathcal{X}}}^{L\times T\times P}$, $R$, $\tau $, $(p_{1}, p_{2}, q_{1}, q_{2})$.
    \OUTPUT $\boldsymbol{\mathcal{X}}^{L\times (T+\tau )\times P}$.
    \STATE \textbf{CP Decomposition:}
    obtain ${\boldsymbol{\mathrm{U}}}_T$ by Eq.(1)
    \STATE
    \textbf{2D-ARMA Prediction:}
    \FOR{$r = 1$ to $R$}
      %\STATE \textbf{Updates}
      \STATE Reshape $\boldsymbol{u}_r$:
      $\boldsymbol{u}_r \in \mathbb{R}^{T}$ to $\boldsymbol{\mathrm{V}}\in \mathbb{R}^{D \times W}$, with $D \times W \geqslant R$
      \FOR{$i=1$ to $\tau$}
        \STATE
        $\boldsymbol{\mathrm{V'}} = f_{2D-ARMA}(\boldsymbol{\mathrm{V}})$ by Eq.(8)
        \STATE
        $\boldsymbol{\mathrm{V}} = \boldsymbol{\mathrm{V'}}$
      \ENDFOR
    \STATE Reshape $\boldsymbol{\mathrm{V'}}$:
    $\boldsymbol{\mathrm{V'}}$ to $\boldsymbol{u'}_r \in \mathbb{R}^{T+\tau} $
    \ENDFOR
    \STATE
    ${\boldsymbol{\mathrm{U}}}_{T+\tau} = [\boldsymbol{u'}_1,\dots,\boldsymbol{u'}_r, \dots, \boldsymbol{u'}_{R}]$
    \STATE \textbf{CP Inverse Reconstruction:}
    obtain $\boldsymbol{\mathcal{X}}^{L\times (T+\tau )\times P}$ by Eq.(1)
    
  \end{algorithmic}
\end{algorithm}

Another challenges is to instantly and efficiently update the prediction result when new data come,  which little research has yielded solutions. The problem is specified in Fig. 4(a). After we obtain the whole predicted passenger flow for tomorrow, when we reach tomorrow 10:00AM for example, the new data will have arrived instead. Then how to update our original prediction dynamically and yield more accurate prediction for the rest of day $T\mathrm{+1}$, especially for next two hours, is a problem. To address this, we propose a lean dynamic tensor decomposition updating method, to alleviate unnecessary workload. The method is demonstrated in Fig. 4(b). 

For the long-term prediction result of day $T\mathrm{+1}$, ${\boldsymbol{\mathcal{X}}}^{L\times 1\times P}$, we propose the following procedure:

\begin{itemize}
\item Decompose the old long-term prediction result into ${\boldsymbol{\mathrm{U}}}_P,{\boldsymbol{\mathrm{U}}}_L$ and ${\boldsymbol{\mathrm{U}}}_T$, with Factor 0 as ${\boldsymbol{\mathrm{U}}}_L$, which needs to be updated when new data arrive.  An illustration is shown in Fig. 4(b).

\item  Assume 30\% new data of it have come, firstly splice the 30\% new data and the rest 70\% long-term prediction together, and update ${\boldsymbol{\mathrm{U'}}}_L$ according to Eq.(9)  \cite{kolda2009tensor}.

\item  Finally, use the updated ${\boldsymbol{\mathrm{U'}}}_L$, and the original ${\boldsymbol{\mathrm{U}}}_T, {\boldsymbol{\mathrm{U}}}_P$, to reconstruct the tensor.
\end{itemize}
 
\begin{equation} \label{GrindEQ__4_} 
\boldsymbol{\mathrm{U'}}_{\mathrm{L}} = \boldsymbol{\mathcal{X'}}_{(0)}({\boldsymbol{\mathrm{U}}}_{\mathrm{T}}\odot{\boldsymbol{\mathrm{U}}}_{\mathrm{P}})({\boldsymbol{\mathrm{U}}}_{\mathrm{T}}^T{\boldsymbol{\mathrm{U}}}_{\mathrm{T}}\ast{\boldsymbol{\mathrm{U}}}_{\mathrm{P}}^T{\boldsymbol{\mathrm{U}}}_{\mathrm{P}})^{\dagger},
\end{equation} 
where $\boldsymbol{\mathcal{X'}}_{(0)}$ is the mode-0 unfolding along L-dimension, $\odot$ is Khatri-Rao product and $^{\dagger}$ is Khatri-Rao product pseudoinverse. 

\begin{figure}[htp]
\subfloat[2-step Tensor Prediction Unable to Update Prediction]{%
  \includegraphics[clip,width=0.95\columnwidth]{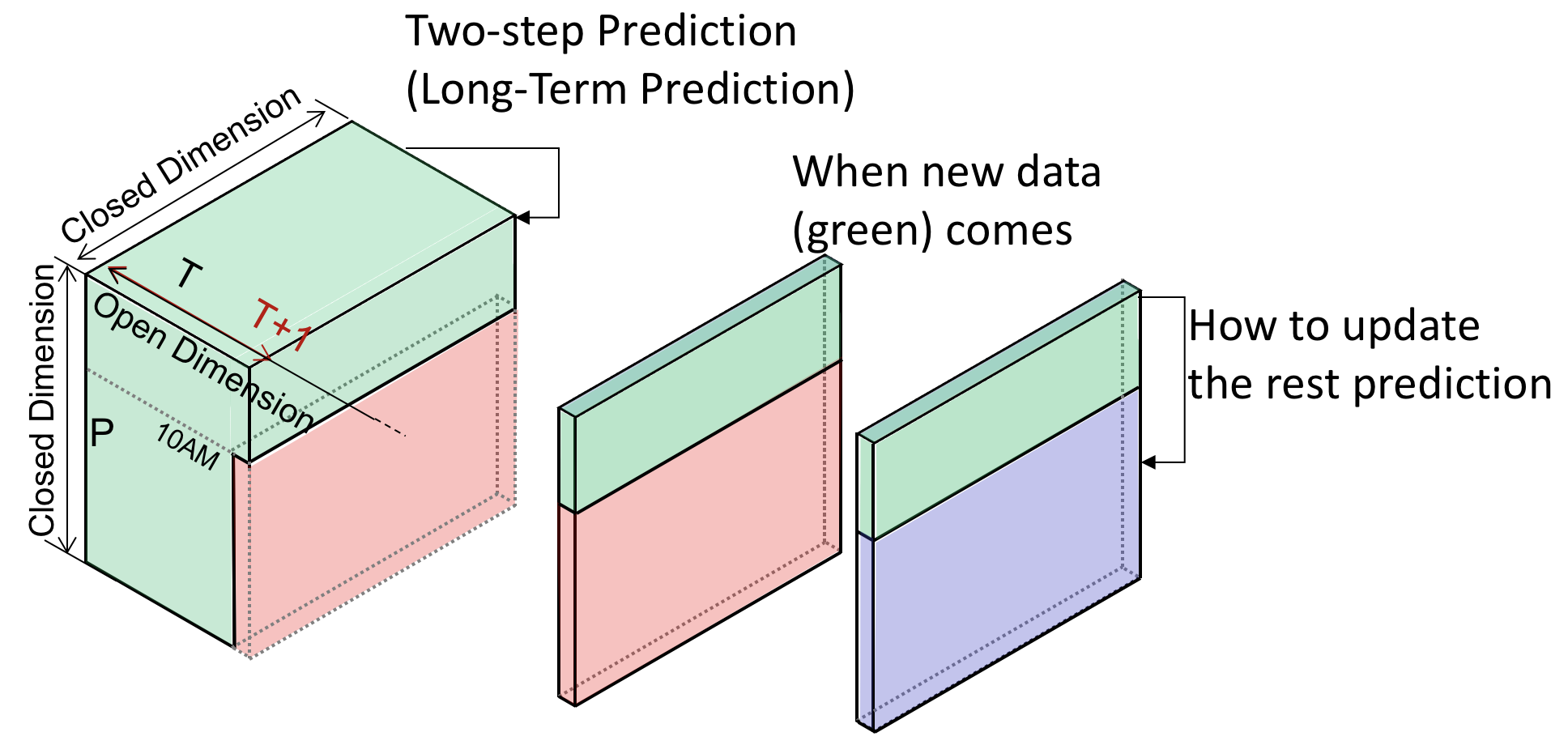}%
}

\subfloat[Lean Dynamic Tensor Decomposition Updating: CP\textit{.fit} is CP composition function, CP\textit{.transform} will be shown in Eq.(9), CP\textit{.inverse} is CP reconstruction function;]{%
  \includegraphics[clip,width=\columnwidth]{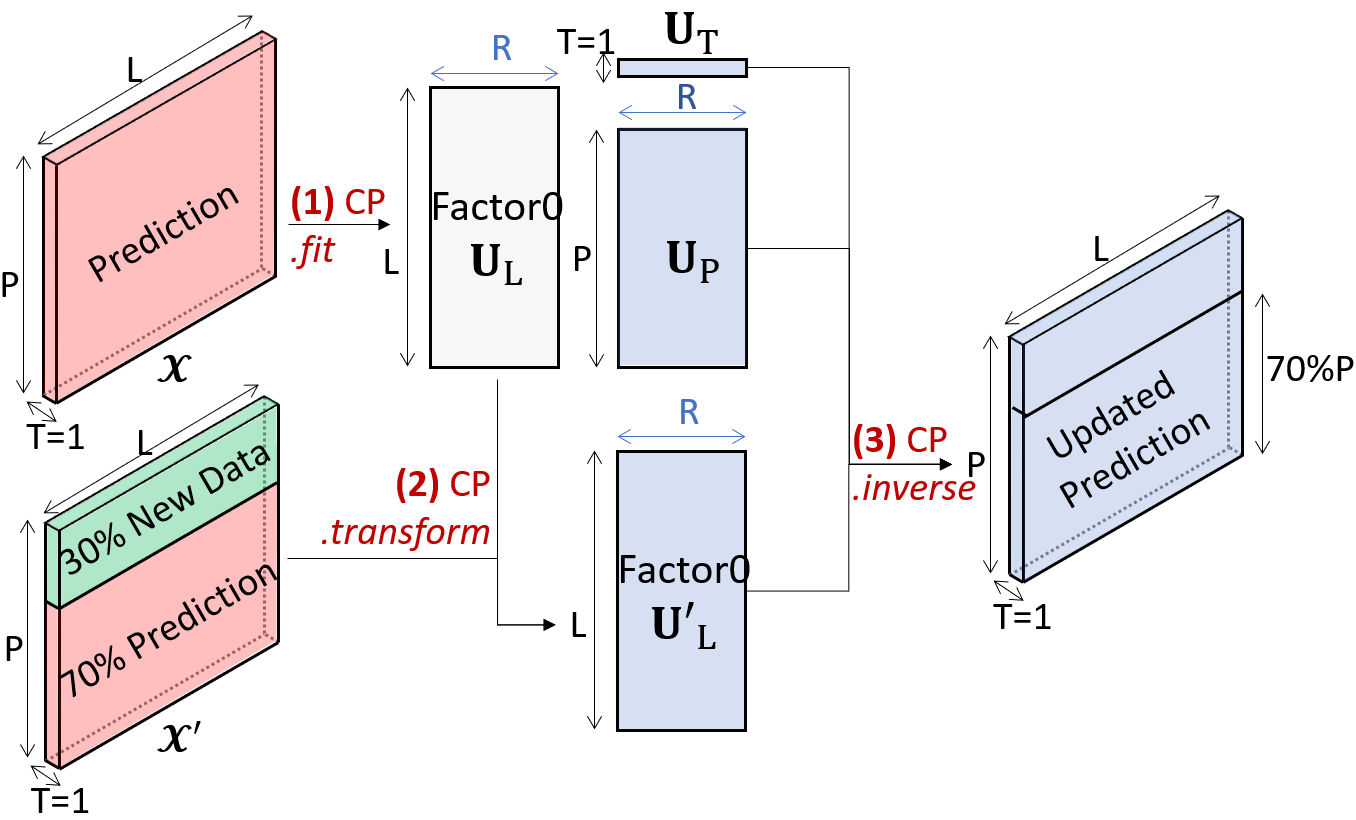}%
}

\caption{Proposed method to Update Prediction}

\end{figure}

In practice, we can combine  the weight $\boldsymbol{\lambda}$ and factor mactrix  0 together and update them when new data arrives. 
%The improvement of updated prediction will be shown in Session 4.

\subsection{ 1-step Tensor Prediction for Short-Term: Tensor Completion}

To adopt tensor completion framework to  prediction, we treat the historical data as observation set (with observation indicator as 1), and the future horizon to be predicted as missing data (with observation indicator as 0). %, shown in Fig. 4.

One thing to be noted is that Tensor completion is not designed for long-term prediction. In particular, here we define two conceptions (as shown in Fig. 4(a)): open dimension and closed dimension. Open dimension is the dimension that keeps increasing as data arrives, e.g. dimension $T$-day here; closed dimension is the one which has fixed maximum length, e.g. dimension $L$-station, and $P$-time point. Tensor completion can be used for missing data imputation along closed dimension (e.g. $P$ and $L$), but not along open dimension ($T$). So it can only predict short-term.

For tensor completion methods, in particular, we follow the Bayesian Low-Rank Tensor Completion (LRTC) framework proposed by Zhao \textit{et al.} \cite{zhao2015bayesian}. 

Denote $\boldsymbol{\mathrm{\Theta }}\mathrm{=}\mathrm{\{}{\boldsymbol{\mathrm{U}}}^{\left(\mathrm{1}\right)}\mathrm{,\dots ,\ }{\boldsymbol{\mathrm{U}}}^{\left(\mathrm{K}\right)}\mathrm{,\ }\boldsymbol{\lambda }\boldsymbol{,\ }\tau \}$ from Eq.(2, 3, 4). After calculating the $log$-joint distribution, and the posterior distribution, the missing data can be estimated after getting $\boldsymbol{\mathrm{\Theta }}$, by:
\begin{equation} \label{GrindEQ__13_} 
P\left({\boldsymbol{\mathcal{Y}}}_{{\boldsymbol{\mathrm{\Omega}}}^c}\boldsymbol{\mathrm{|}}{\boldsymbol{\mathcal{Y}}}_{\boldsymbol{\mathrm{\Omega}}}\right)=\int{P\left({\boldsymbol{\mathcal{Y}}}_{{\boldsymbol{\mathrm{\Omega}}}^c}|\boldsymbol{\mathrm{\Theta }}\right)}P\left(\boldsymbol{\mathrm{\Theta }}\boldsymbol{\mathrm{|}}{\boldsymbol{\mathcal{Y}}}_{\boldsymbol{\mathrm{\Omega}}}\right)\mathrm{\ d}\boldsymbol{\mathrm{\Theta }}.
\end{equation} 

However, this method still demonstrates some drawbacks as noted in \cite{yuan2018high}.  In particular, when the tensor data violate the innate low-rank structure, such as in our case that the spatial information of URT tensor data is diverse from stations to stations, the low-rank assumption along the spatial dimension cannot hold tenably. Consequently, it is far from enough to assume the spatial prior as low-rank. This most likely oversimplifies the original data structure. 

\begin{figure}[htp]
\centering
%\framebox{\parbox{3in}{We suggest that you use a text box to insert a graphic (which is ideally a 300 dpi TIFF or EPS file, with all fonts embedded) because, in an document, this method is somewhat more stable than directly inserting a picture.}}
\includegraphics[width=0.95\columnwidth]{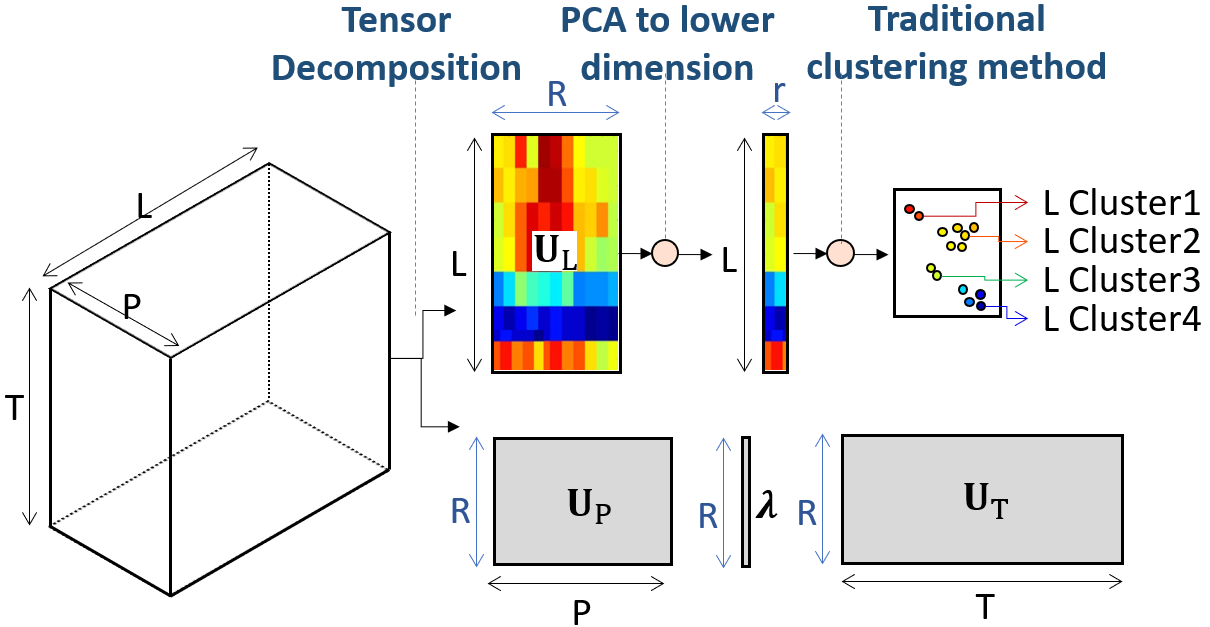} % Reduce the figure size so that it is slightly narrower than the column. Don't use precise values for figure width.This setup will avoid overfull boxes. 
\caption{Tensor Clustering based on Tensor Decomposition}
\label{fig7}
\end{figure}

To solve the problem, before the Bayesian LRTC, we proposed to use tensor clustering first \cite{yu2019coupled,sun2019dynamic} to classify the tensor samples into several classes, with highly similar samples within a same cluster. Tensor clustering method is shown in Fig. 5 with the following steps:

\begin{itemize}
\item  Conduct Tensor decomposition to obtain location model matrix.

\item  Implement Principal Component Analysis to further reduce the dimension.

\item  Cluster based on a particular clustering method, such as K-mean and Hierarchical method. 
\end{itemize}

Thus different URT stations can be divided into several clusters, and Bayesian LRTC can be conducted within each cluster since the homogeneity within cluster can guarantee its performance.

\section{Experiments}

According to what we have discussed in Session III, our UTR passenger flow tensor data is ${\boldsymbol{\mathcal{X}}}^{120\times 59\times 247}$, representing 120 stations, over the past 59 days, with each day 247 sampling points. We set the first 50 days as known historical data and the last 9 days as data to be predicted. 

\begin{figure}[htp]
\centering
%\framebox{\parbox{3in}{We suggest that you use a text box to insert a graphic (which is ideally a 300 dpi TIFF or EPS file, with all fonts embedded) because, in an document, this method is somewhat more stable than directly inserting a picture.}}
\includegraphics[width=0.95\columnwidth]{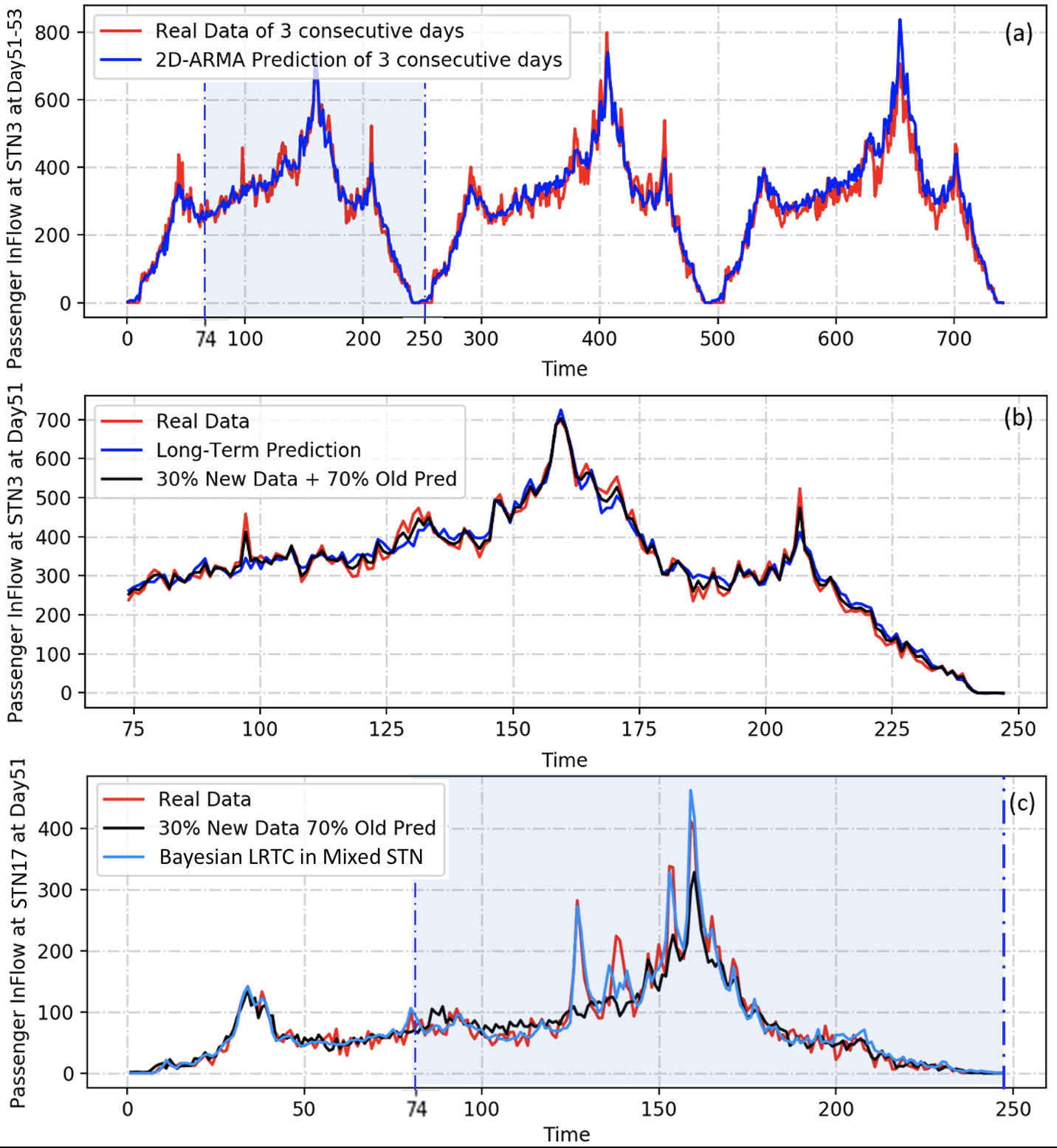} % Reduce the figure size so that it is slightly narrower than the column. Don't use precise values for figure width.This setup will avoid overfull boxes. 
\caption{(a) Inflow Profile Long-Term Prediction for STN3 by 2-step 2D-ARMA prediction; (b) Prediction Improvement for STN3 by involving 30\% new data; (c) Inflow Profile Short-Term Prediction for STN17 by 1-step prediction}
\label{fig7}
\end{figure}

\begin{table}[h]
\caption{2-STEP PREDICTION COMPARISON (RES)}
\begin{center}
\begin{tabular}{|p{1.2cm}||p{1.8cm}||p{1.8cm}||p{1.8cm}|}
\hline
Randomly Selected Station Code & 1D ARIMA Tensor Prediction & 2D ARMA Tensor Prediction & Relative improvement (\%)\\
\hline
51 & 0.1500 & 0.1066 & \textbf{28.90}\\
56 & 0.1043 & 0.0759 & \textbf{27.21}\\
38 & 0.0849 & 0.0638 & \textbf{24.91}\\
87 & 0.1400 & 0.1087 & \textbf{22.39}\\
54 & 0.0939 & 0.0739 & \textbf{21.33}\\
84 & 0.1605 & 0.1271 & \textbf{20.81}\\
11 & 0.0962 & 0.0767 & \textbf{20.29}\\
16 & 0.0871 & 0.0703 & \textbf{19.30}\\
37 & 0.0982 & 0.0811 & \textbf{17.38}\\
65 & 0.1247 & 0.1060 & \textbf{15.00}\\
%57 & 0.1783 & 0.1546 & \textbf{13.30}\\
%64 & 0.0917 & 0.0807 & \textbf{11.98}\\
\hline
\end{tabular}
\end{center}
\end{table}

\subsection{Proposed 2-step Tensor Prediction Result}
For the tensor ${\boldsymbol{\mathcal{X}}}^{120\times 50\times 247}$, we conduct CP decomposition. The rank is chosen as 50 by cross validation, achieving both satisfactory reconstruction and simplicity. For each rank, a 2D ARMA model is constructed and used to predict the coming 9-day-ahead passenger flow. The results of the first 3 days long-term prediction are shown in Fig. 6(a). In our proposed method, to capture the weekly pattern of past 2 weeks and the daily pattern of past 2 days, we set $p_{1}=2, p_{2}=2$, which introduces 8 time-lag components into model. For fair comparison, the baseline is chosen as the traditional "Tensor Decomposition + 1D ARIMA" model, with the same model complexity time-lag equal to 8. As shown in Table I, our proposed model can achieve almost 20\% improvements, evaluated by relative residual (RES). The overall improvement benefits from the advantage that more strongly correlated temporal patterns have been considered in our model. Note that the scale of improvement varies. This is because some stations may also have strong temporal correlation with time-lag(3), (4) etc., which yet our model ignores.

\begin{table}[htp]
\caption{IMPROVEMENT ON LONG-TERM PREDICTION (RES)}
\begin{center}
\begin{tabular}{|p{0.9cm}||p{1.8cm}||p{2cm}||p{1.8cm}|}
\hline
[t, t+5] & Long-Term Prediction & Updated after 30\% new data & Relative improvement (\%)\\
\hline
t =75 & 0.9221 & \textbf{0.8436} & \textbf{8.520}\\
t =80 & 0.5833 & \textbf{0.3920} & \textbf{32.79}\\
t =85 & 0.7136 & \textbf{0.6258} & \textbf{12.29}\\
t =90 & 1.2381 & \textbf{1.2042} & \textbf{2.735}\\
t =95 & 1.2543 & \textbf{1.1387} & \textbf{9.215}\\
t =100 & 1.0423 & \textbf{0.9729} & \textbf{6.659}\\
t =105 & 0.1153 & \textbf{0.0743} & \textbf{35.57}\\
t =110 & 0.3452 & \textbf{0.3379} & \textbf{2.113}\\
t =115 & 0.4022 & \textbf{0.3163} & \textbf{21.36}\\
t =120 & 0.2292 & 0.3084 & -34.54\\
t =125 & 0.6245 & 0.6865 & -9.936\\
%t =130 & 0.2383 & \textbf{0.2467} & \textbf{16.16}\\
\hline
\end{tabular}
\end{center}
\end{table}

When the first 30\% new data (in Fig. 6(a) until time stamp $t=74$) have arrived, the prediction for the rest 70\% of that day (In Fig. 6(a) highlighted in blue block, from time stamp $t=75$ to $t=247$) needs to be updated instantly. By using proposed lean dynamic updating, the rest 70\% has been recalculated as shown in Fig. 6(b). It is clear to observe that the updated prediction is significantly improved with smaller distance to the real value, especially around the local peak time. To further check the improvement of the rest 70\%, RES is calculated from $t=74$ for every 25 minutes (i.e., 5 time stamps). According to Table II, after involving the first 30\% of new data, there is an obvious improvement by around 20\% (highlighted in boldface) over the following 3 hours (from $t=75$ to $t=115$). After then, the relative improvement based on short-term updating becomes less efficient, with the long-term prediction still being preferred.

\subsection{Proposed Bayesian LRTC Result}

For the Bayesian LRTC, some stations from mixed clusters have been randomly picked, and the data since $t=74$ (around 10AM) of last day are to be predicted. The prediction result is shown in Fig. 6(c). The Bayesian LRTC can reduce RES by 29\% for station 17  in Fig. 6(c), with some other stations achieving around 10\% to 30\% less residual as shown in Table III. The good performance for tensor completion in short-term prediction is quite satisfactory compared with the proposed Lean Dynamic Updating. However, according to Table III, it is to be noted that this improvement (highlighted in boldface) is not universal, with station 84, 55, 51 and 87 having worse prediction. This is because these four stations have quite unique and distinct passenger flow pattern and the low rank assumption no longer holds.

\begin{table}[htp]
\caption{PREDICTION COMPARISON FOR MIXED-CLUSTER (RES)}
\begin{center}
\begin{tabular}{|p{1.2cm}||p{1.8cm}||p{1.8cm}||p{1.8cm}|}
\hline
Station Code (Mixed Cluster) & Updated after 30\% new data & Bayesian LRTC & Relative improvement (\%)\\
\hline
17 & 0.1165 & 0.0821 & \textbf{29.59}\\
56 & 0.0885 & 0.0649 & \textbf{26.66}\\
%57 & 0.1450 & 0.1216 & \textbf{16.18}\\
54 & 0.0833 & 0.0704 & \textbf{15.38}\\
38 & 0.1775 & 0.1521 & \textbf{14.31}\\
%64 & 0.1565 & 0.1353 & \textbf{13.55}\\
35 & 0.1020 & 0.0909 & \textbf{10.89}\\
11 & 0.1416 & 0.1326 & \textbf{6.38}\\
\underline{84} & \underline{0.0632} & \underline{0.0691} & \underline{-9.38}\\
55 & 0.1240 & 0.1366 & -10.17\\
51 & 0.1090 & 0.1279 & -17.38\\
\underline{87} & \underline{0.1205} & \underline{0.1609} & \underline{-33.60}\\
\hline
\end{tabular}
\end{center}
\end{table}

This can be solved by conducting Bayesian LRTC within a same station cluster. We use the Hierarchical Clustering with Agglomerative Method, where the distance is defined as Group Average Euclidean. The clustering result reflects two types of spatial dependencies observed in our data: Law of Geography and contextual similarity. In other words, if two stations are geographically close or functional similar, they are in the same cluster. For example, Fig. 7 shows the land-use information of three selected stations (Station 73, 80 and 84) in one cluster. Though Station 73 and 84 are far to each other, they share the same land-use pattern (with red dominant, mix use in grey and brown). Though Station 84 and 80 are quite different in land-use (Station 84 is dominated with red, Station 80 is dominated with green), they are geographically close (since the stations are indexed consecutively, two stations with close codes indicates they are physically close.)

\begin{figure}[htp]
\centering
%\framebox{\parbox{3in}{We suggest that you use a text box to insert a graphic (which is ideally a 300 dpi TIFF or EPS file, with all fonts embedded) because, in an document, this method is somewhat more stable than directly inserting a picture.}}
\includegraphics[width=0.95\columnwidth]{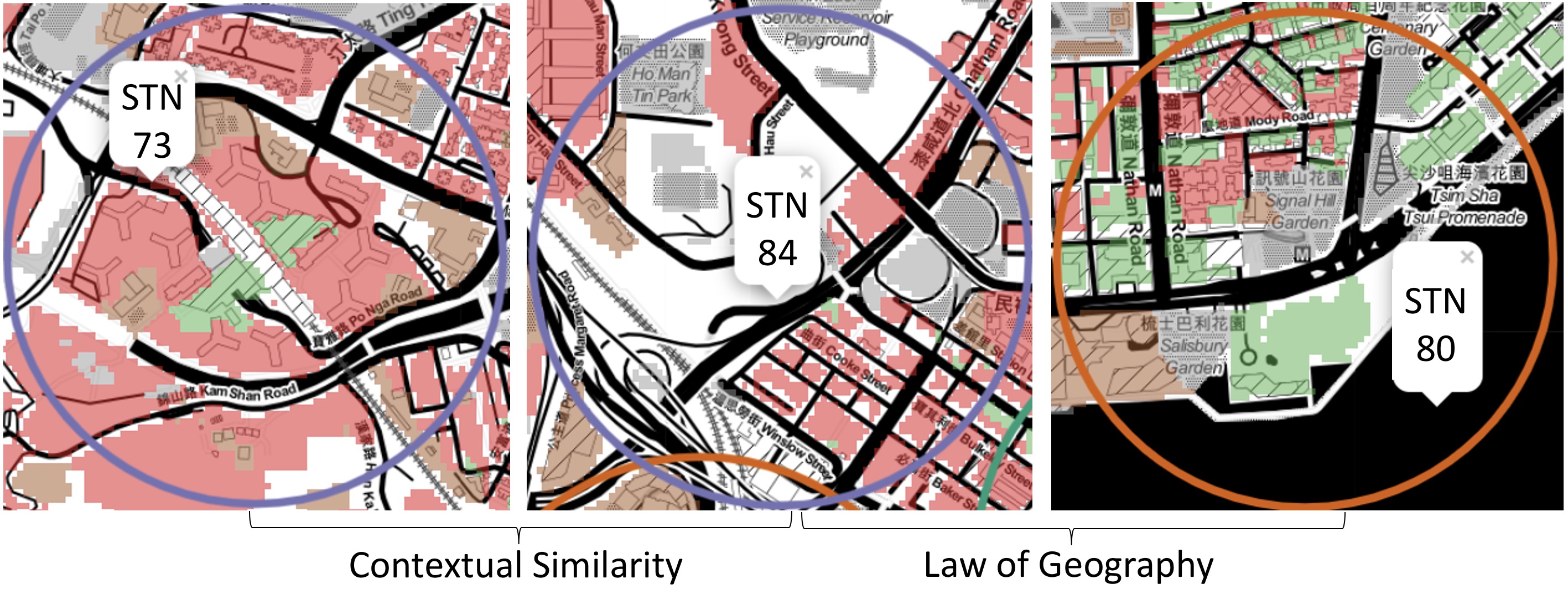} % Reduce the figure size so that it is slightly narrower than the column. Don't use precise values for figure width.This setup will avoid overfull boxes. 
\caption{Land-use Information of Selected Stations (Different Color Pixel Denotes Different Land-use)}
\label{fig9}
\end{figure}

By selecting stations within this cluster, we compared the Bayesian LRTC within one cluster with 2-step Prediction updated with 30\% new data, and the result is shown in Table IV . 

All the stations' predictions have been improved (with improvement highlighted in boldface), with the majority improved by 20\%. Most importantly, even for the same Station 84 and 87 in two cases(highlighted with underline in Table III and IV), conducting Bayesian LRTC within one cluster can improve prediction by 30\% to 40\% .

\begin{table}[htp]
\caption{IMPROVEMENT ON LRTC BY SAME CLUSTER (RES)}
\begin{center}
\begin{tabular}{|p{1.2cm}||p{1.8cm}||p{1.8cm}||p{1.8cm}|}
\hline
Station Code (Same Cluster) & Updated after 30\% new data & Bayesian LRTC within a cluster & Relative improvement (\%)\\
\hline
82 & 0.1149 & 0.0673 & \textbf{41.43}\\
77 & 0.1775 & 0.1133 & \textbf{36.18}\\
79 & 0.2112 & 0.1349 & \textbf{36.10}\\
71 & 0.1109 & 0.0766 & \textbf{30.89}\\
\underline{84} & \underline{0.0632} & \underline{0.0489} & \underline{\textbf{22.59}}\\
75 & 0.1593 & 0.1288 & \textbf{19.15}\\
74 & 0.1642 & 0.1409 & \textbf{14.19}\\
73 & 0.1773 & 0.1567 & \textbf{11.61}\\
80 & 0.1187 & 0.1087 & \textbf{8.41}\\
%72 & 0.1782 & 0.1635 & \textbf{8.24}\\
%76 & 0.1298 & 0.1204 & \textbf{7.26}\\
\underline{87} & \underline{0.1205} & \underline{0.1146} & \underline{\textbf{4.92}}\\
\hline
\end{tabular}
\end{center}
\end{table}

\section*{Conclusion}

In this paper, we focused on both Long-term and Short-Term URT passenger flow prediction. 

Our proposed 2-step Tensor Prediction based on Tensor decomposition and time-series model can predict both long-term and short-term.  In particular, the "CP Decomposition + 2D ARMA model" can achieve satisfactory long-term prediction, and the lean tensor decomposition updating method can update short-term prediction after receiving new data.

Our proposed 1-step Tensor Prediction based on Bayesian Low Rank Tensor Completion can only predict short-term, but with better performance than lean dynamic tensor decomposition updating. To solve its the innate drawback of low-rank assumption that results in oversimplification, a tensor cluster technique is first implemented and then Tensor Completion is conducted for each cluster respectively.

\addtolength{\textheight}{-9.5cm}   % This command serves to balance the column lengths

%%%%%%%%%%%%%%%%%%%%%%%%%%%%%%%%%%%%%%%%%%%%%%%%%%%%%%%%%%%%%%%%%%%%%%%%%%%%%%%%

%%%%%%%%%%%%%%%%%%%%%%%%%%%%%%%%%%%%%%%%%%%%%%%%%%%%%%%%%%%%%%%%%%%%%%%%%%%%%%%%

%%%%%%%%%%%%%%%%%%%%%%%%%%%%%%%%%%%%%%%%%%%%%%%%%%%%%%%%%%%%%%%%%%%%%%%%%%%%%%%%
\section*{ACKNOWLEDGMENT}

This research is supported by Hong Kong MTR Co. with grant numbers RGC GRF 16203917, 16201718 and NSFC 71931006. We also give special thanks to Dr. Qibin Zhao for sharing the scripts of their methods.

%%%%%%%%%%%%%%%%%%%%%%%%%%%%%%%%%%%%%%%%%%%%%%%%%%%%%%%%%%%%%%%%%%%%%%%%%%%%%%%%
%\section*{APPENDIX}

%%%%%%%%%%%%%%%%%%%%%%%%%%%%%%%%%%%%%%%%%%%%%%%%%%%%%%%%%%%%%%%%%%%%%%%%%%%%%%%%

%\bibliographystyle{./IEEEtran}
%\bibliography{./IEEEabrv,./reference}
\bibliographystyle{IEEEtran}
\bibliography{reference.bib}

\end{document}